\documentclass[letterpaper, 10 pt, conference]{ieeeconf} 

\usepackage{graphicx}
\usepackage{subcaption}
\usepackage[ruled,vlined]{algorithm2e}
\usepackage{lineno}
\usepackage{hyperref}
\usepackage{flushend}

\IEEEoverridecommandlockouts                               
\title{\LARGE \bf
Object Goal Navigation Based on Semantics and RGB Ego View 
}

\author{Snehasis Banerjee$^{1,2}$, Brojeshwar Bhowmick$^{1}$ and Ruddra Dev Roychoudhury$^{1}$
\thanks{$^{1}$TCS Research, Tata Consultancy Services, Kolkata
        {\tt\small Primary Author - $^{2}$ snehasis.banerjee@tcs.com}}%
}
\begin{document}

\maketitle
\thispagestyle{empty}
\pagestyle{empty}

\begin{abstract}
This paper presents an architecture and methodology to empower a service robot to navigate an indoor environment with semantic decision making, given RGB ego view. This method leverages the knowledge of robot's actuation capability and that of scenes, objects and their relations -- represented in a semantic form. The robot navigates based on GeoSem map - a relational combination of geometric and semantic map. The goal given to the robot is to find an object in a unknown environment with no navigational map and only egocentric RGB camera perception. The approach is tested both on a simulation environment and real life indoor settings. The presented approach was found to outperform human users in gamified evaluations with respect to average completion time.

\end{abstract}


\section{Introduction}
Recent advances in AI and Robotics has led to the emergence of service robots capable of complex navigation tasks, which has two well established paradigms -- (a) geometric reconstruction and path-planning based approaches (b) end-to-end learning-based methods. To navigate an environment, the robot needs to sense the scene in some way. In this tasks, camera sensor has been the primary choice for perception in navigation tasks along with relevant distance measure sensors such as echo, infra-red, lidar. However, among the popular sensors, RGB camera has a lower cost and higher availability. In fact to make service robots more affordable and mainstream, a system using only RGB perception and relying on software intelligence to enable navigation capability is highly desired. This paper is an effort in that line to use only RGB perception in a wheeled robot to enable smart navigation using fast semantic inferences.

Humans are generally very good at the task of navigation. If a person is asked to find an object in a unknown scene, his (or her) decision making will be based on visual cues in current scene. Inspired by aforementioned human intuition, in this work, the robot takes navigation decision based on an amalgation of current zone's probability of having an object, visible objects that are closely related to the target object, visible occlusions that may hide the object along with other factors like free space, obstacles and risky or restricted areas.

\subsection{Problem Description} 
The task is to navigate in an unmapped indoor environment from a starting point to a specified object location and the task is marked as complete if the object is visible to a practical extent. The task needs to be completed based on RGB perception of egocentric view of onboard robot camera.

The main contributions of the paper are as follows:\\
(i) object goal navigation algorithm using `GeoSem' map and semantic knowledge, with RGB ego view image as input.\\
(ii) a system architecture to enable semantic navigation.\\
(iii) successful functional evaluation of proposed methodology in both simulated and real world indoor environments.

\section{Related Work}
Although there has been a considerable amount of work on the general object goal navigation problem, there has been limited work on searching an object type just based on egocentric RGB camera mounted on a robot in indoor settings using semantics. A number of works \cite{gireesh2022object} uses deep neural network for object-centric navigation policies. The closest work on Semantic Visual Navigation \cite{kiran2022spatial} uses scene priors to navigate to known as well as unknown objects in limited settings using Graph Convolutional Networks (GCNs). The aforementioned work has the following limitations: (a) no concrete decision model when two or more objects are in same frame (b) actual motion planning is not formulated (c) No testing on real life scenario (d) Deep Reinforcement Learning framework requires significant amount of training to learn action policies. Our approach addresses the aforementioned limitations in the sense that it tackles multiple objects in single frame by design; path planning methods are integrated in the decision making process; it has been tested in limited but real world scenario; does not require extensive training; and is based on knowledge-based decision making. 
\begin{figure}[t]
	\centering
	\includegraphics[width=0.8\linewidth]{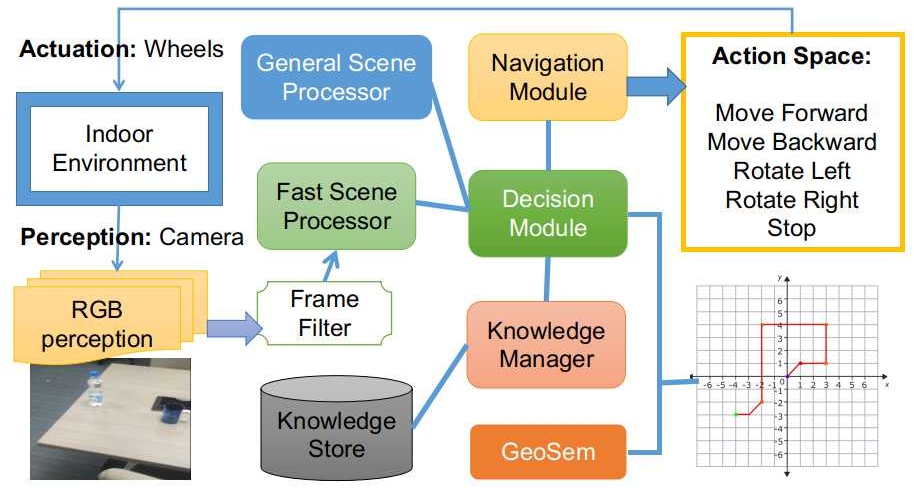}
	\caption{System Architecture for Semantic Navigation}
	\label{fig:arch}
\end{figure}
\section{System Architecture}
Fig. \ref{fig:arch} presents the system architecture. For a wheeled robot, the Action Space is limited to the movements in forward direction or backward; and rotations towards left or right with respect to ego view; and stop. The system takes in perception from RGB camera mounted in first person view of robot and gives out an action control command for the robot to move. The next RGB view is the only feedback post actuation that the robot can get. However, as the action space, in this case, by assumption, is of discreet nature, the robot is capable of drawing a map of its movements internally as discussed in Section IV.B. This is susceptible to drift errors in real life deployment, however they can be corrected by loop closure algorithms \cite{tsintotas2022revisiting}. The robot waits in `stop' state until current scene is processed. Generally, scene processing happens continuously at 30 fps. But, the system is incapable of handling such a fast rate of image processing and consequent decision making, due to algorithm complexity related time restrictions. Hence, a Filter module is kept to handle the inflow of images and drop frames in order to match processing speed. Although Mask-RCNN \cite{he2017mask} based object detection and segmentation has been established to perform very well, however in robotic scenario (soft real time system) its performance speed of  $\sim$ 2 fps in standard hardware does not conform to the affordibility aspect. Hence, we have chosen a fast scene processor namely YOLO v.3 \cite{redmon2018yolov3} for this task after comparing performance with various scene processors \cite{han2018advanced} in indoor settings. However, YOLO has problems in detecting two or more objects lying in same the grid box. Hence, if YOLO is unable to detect objects in current scene, the more robust General Scene Processor (Mask-RCNN based module) is invoked. From experiments, it was found that YOLO was successful in detecting objects most of the time. The Knowledge Manager is concerned with the task of accessing the Knowledge Store (ontologies and facts) and answering queries from Decision module. It will also update the knowledge store with scene relations learned over time which is added as a weighted score relative to number of scenes explored. Based on internal inferences drawn on processed scene, the module updates the GeoSem map (discussed in Section IV.B) and issues next movement command to Navigation module. Navigation module is tasked with maintaining a synchronized communication with the robot's actuation (motor movements) and notifying the Decision module once a movement is complete.
\section{Methodology}
\subsection{Knowledge Representation}
To represent semantic knowledge, we have chosen semantic web technology \cite{gandon2018survey}, due to its standards and extension capacity. Categorical and set theoritic relations in robotics are represented in OWL\footnote{https://www.w3.org/OWL/} \cite{tenorth2017representations} and model instances in RDF\footnote{https://www.w3.org/RDF/} files. Logical queries based on SPARQL\footnote{https://www.w3.org/TR/rdf-sparql-query/} enable reasoning on the instance relations. These relations are learned by processing scenes from Visual Genome dataset \cite{krishna2017visual} based on frequency statistics. 

Some of the semantic relations are enlisted here:

(1) locatedAt(object, zone): this denotes the probability of an object (say TV) to be located in a zone (say video conference room). Zones related to risk or restrictions (say security room) can also be modeled in this way.

(2) coLocatedWith(object$_1$, object$_2$) : this symmetric relation denotes the probability of an object (say `bottle') to be co-located with another object (say `cup'). The longer the edge length between objects in this generated graph from scene processing, the less is the chance of co-location relation. Assuming $\beta$ as a parameter dependent on graph density, the relation probability \textit{RP} =\\\\
$\frac{\sum_{i=1}^{n} \beta \cdot \frac{1}{n} \cdot distinct\_colocation\_probability(object_i , object_n)}{\sum distinct\_edges}$\\

(3) locatedOnTopOf(object$_1$, object$_2$): this spatial relation denotes if an object$_1$ (say `cup') can reside on top of another object$_2$ (say `table'). Its reverse relation is `locatedBelow'.

(4) occlusionBy(object$_1$, object$_2$): this relation denotes the probability that an object$_1$ can get hidden or occluded by another object$_2$ --- this is based on the relative average dimensions of the objects. These objects should have a high `coLocatedWith' or co-occurence probability.

\begin{algorithm}[t]
	\SetAlgoLined
	\KwResult{Target object becomes visible in ego view}
	\textbf{Parameters}:\\ 
	image $\leftarrow$ RGB camera egocentric image stream\;
	actuation $\leftarrow$ movement commands to robot wheels\;
	knowledge $\leftarrow$ link to semantic knowledge store\;
	geosem $\leftarrow$ geometric semantic map representation\;
	software $\leftarrow$ link to software modules;\\
	target $\leftarrow$ target object to search (user instruction)\;
	sa $\leftarrow$ area of scene (example: Left, Middle, Right)\;
	\While{ target != found or geosem.Scan != full }{
		Wait for actuation completion\;
		\If{ geosem.landmarkScore \textbf{Is} Low }{
			actuation $\leftarrow$ geosem.past\_actuation\;
			continue\;
		}
		sa, objects $\leftarrow$ software.objectDetect(image)\;
		knowledge.updateRelations(objects, sa)\;
		\If{`restricted objects' \textbf{Not In} sa[i to n]}{
			sa $\leftarrow$ sa[ keep matched indices ]\;
		}
		\If{`obstacle objects' \textbf{Not In} sa[i to n]}{
			sa $\leftarrow$ sa[ keep matched indices ]\;
		}
		\uIf{ knowledge.ZoneRelation(sa) \textbf{Is} Low }{
			\If{`opening' or `doors' \textbf{In} sa}{
				actuation $\leftarrow$ sa[found area]\;
				continue\;
			}
		}
		\uElseIf{ `relational objects' \textbf{in} sa }{
			For each sa:\\
			c $\leftarrow$ centroid of knowledge.RP(object, target) * software.confidence\_score(object)\;
			actuation $\leftarrow$ sa[ max (c) ]\;
			\uElseIf{ `free space' \textbf{in} sa }{
				actuation $\leftarrow$ sa[ free space ]\;
			}	
		}\uElse{
			actuation $\leftarrow$ geosem.LastStep\;
		}
		geosem.Update(actuation, objects, sa)\;
	}
	\If{geosem.Scan == full}{
		print `Target Not Found.;	actuation $\leftarrow$ STOP\;
	}
	\caption{Object Goal Navigation using Semantics}
	
\end{algorithm}

\subsection{GeoSem Map}
In this section, we introduce the concept of GeoSem Map, which builds a semantically rich map based on frontal RGB perception combined with geometrical movement monitoring of the robot in defined steps. It is based on a eucledian co-ordinate system, where robot is placed initially at the origin. If the actuation capability of robot is limited to certain action steps (like move forward by `M' metre or rotate left by `D' degrees), then the robot's location in the 2-D map can be referred by distinct (x,y) co-ordinates with an angle $\theta$ as direction of camera view. If the robot's action space is discreet like having fixed movement capacity, then the robot should rotate at 90$^0$ angles. Otherwise, the robot's rotation angle should be 45$^0$ each time, to help it navigate the map in grid forms of fixed square sides and diagonals. If an obstacle is encountered in ego view of next trajectory direction, it will avoid the obstacle by going around it and restoring its trajectory by taking help of this map. After each successful movement in action space, the scene is analyzed to find relational objects to determine its next move. While the robot is completing its action move, scene processing should happen only to handle dynamic obstacles, thereby saving computation time. We have kept handling dynamic obstacles out of scope. Each location step stores a tuple comprising of -- the relative direction of ego view, objects in scene with confidence score and a calculated landmark score for that point. Ideally as the robot moves forward, the relative landmark score should keep on increasing until target object is found. A diminshing landmark score signifies that the robot is going in wrong direction. Landmark score =\\\\
$\frac{\sum prob(object, target) }{\sum objects} \cdot prob(zone, target) \cdot \alpha \cdot \frac{\sum | rotation^0 |}{360^0}$\\

Landmark score is based on the probability of the target object being found in the scene as well as the combined probabilities of objects located in scene and their relations to target object (accessed from ontology). Whether a point in scene has been explored in totality is understood from rotational moves at that zone. $\alpha$ denotes a scaling factor specific to environmental settings.
\begin{figure}[h]
	\centering
	\includegraphics[width=0.7\linewidth]{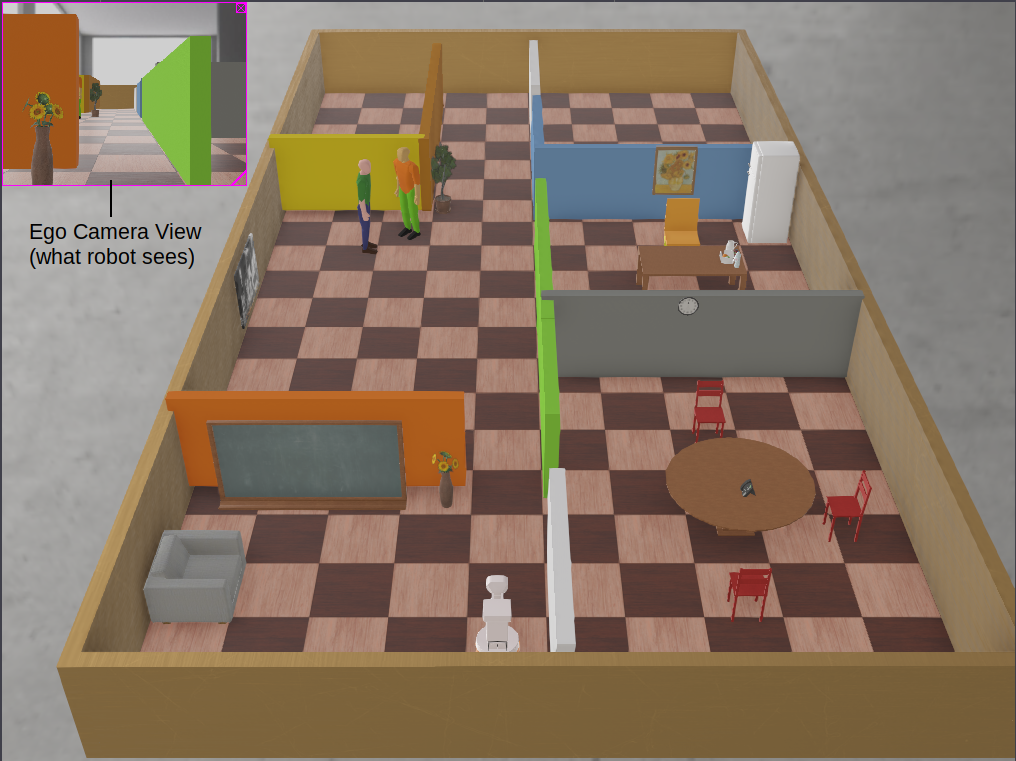}
	\caption{Indoor Environment simulation (gamified)}
	\label{fig:webots}
\end{figure}
Contrary to SLAM \cite{saputra2018visual} based mapping, the robot's primary task here is not to map the environment, but to reach the target. In this case, the objective of mapping is to remember landmark points (with respect to target finding) traversed so that it can backtrack to an earlier location if current trajectory is not successful, or can shorten the backtrack exploration if the robot finds itself close to the previously traversed path point(s).
%
\subsection{Scene Processing}
Application of object detection algorithm (on RGB perception) in a scene will reveal four categories of objects: (a) Target object (b) relational objects (c) obstacles (d) generic scene object extensions (like door, opening, wall, floor, ceiling). There can be overlap in the object categories. As an example, floor denotes free space, walls are static obstacles. Walls on both sides with floor in between means a `Passage'. If the object relations in scene relative to a zone (say room) is low, the robot should move to a different scene, else it will explore the scene by further rotation and consequential exploration. In the same scene, if occlusion objects exist, then, the robot should search for free space to go around identified occlusion in an attempt to bring the object in view. The ego view scene (measured by image height and weight) is broken into polygonal segments. The philosophy is that the robot should move towards the polygon segment having one or more objects with maximum probable relation to target object (refer to relation probability \textit{RP} in Section IV.A). The decision to move towards a polygon segment is based on the centroid having maximum summation of \textit{RP} by combination of corresponding objects' location in the scene. 

\subsection{Navigation Algorithm}
The navigation algorithm combines approaches of visual semantics as well as geometry based decision making based on estimated robot actuation. With the initial view of the robot at start location (namely ego scene ES$_1$) the following steps are carried out: scene is processed for object detection and corresponding GeoSem map updation. If target object is luckily in ES$_1$, then the goal is achieved. If a door or some opening into another zone (say room) is observed nearby (based on object category detection), the robot moves in free space close to that zone for doing a rotation scan. If no openings are found, the robot does a full 360$^{0}$ rotational scan for scene analysis. Then it determines which way to move based on relational decision making and occlusion estimates. The robot will move towards the zone that has highest chance of finding the object. If an obstacle falls in the way it will bypass it following a `Manhattan block' grid motion pattern. The robot can backtrack to a point in path (with next high probability of target object finding) when its current trajectory path has reached a dead end. This process is repeated until the robot can view the object or the exhaustive search upto a limit is over. The knowledge of the metric dimension of the robot's body helps in approximating spaces that it should avoid when going through doors and passages, or avoiding obstacles.

\section{Evaluation}
An indoor environment (refer Fig. \ref{fig:webots})
was created using Webots\footnote{https://cyberbotics.com/} and a robot model `Tiago Iron'\footnote{https://tiago.pal-robotics.com/}, was selected to simulate wheeled navigation. The robot was set at an initial position away from the target object and issued a goal to find target object -- `orange'. The movement commands come from the discreet set of actions - rotate left, rotate right, go back, go forward and stop. 30 human users  (IT professionals aged between 22 to 38 years) were given keyboard controls of the action space and were asked to play the game of navigating to target object from initial scene in this Webots custom scene setup. The start location and ego camera frame was same for human users and robot. The simulation environment was a black box for the robot with input as current ego view image and output as next actuation move. The average time for successful human navigation was around 6 minutes, where as minimum and maximum time taken where 1 minute 23 seconds and 13 minutes 12 seconds respectively. For the same task, the robot took a time around 2 minutes. The task was marked complete when object detector algorithm could detect target object from ego view with good confidence.

\begin{figure}[h]
	\centering
	\includegraphics[width=0.8\linewidth]{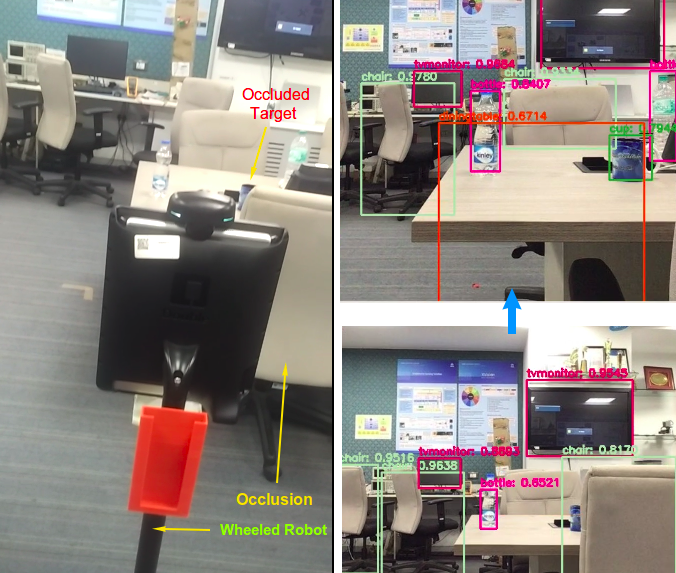}
	\caption{Real world scenario: robot moving Forward to visualize Target object `cup' occluded behind object `chair' with reference to relational objects in scene: `bottle' and `table'}
	\label{fig:demo}
\end{figure}

Some of the observations from the gamified experiment (Table 1) are: (a) most people preferred a strategy similar to breadth first search exploration (b) first person view (egocentric camera view) was found tough in terms of entering new zones and avoiding collision (c) most people ignored looking at objects at the initial back view of the robot and moved forward from initial view. (d) users conversant with first person view games performed well.
In another experiment, in a real world office environment, the task given to the robot was to find target object `cup' (refer Fig. \ref{fig:demo}). The software was deployed as a ROS software node in a wheeled robot Double2 \footnote{https://www.doublerobotics.com/double2.html}. The robot finds related objects `bottle' and `table' in the scene and infers that there is a high co-location probability of `bottle' as well as `table' with `cup'. The spatial relational probability of `cup' on top of `table' was also found high. Also, in the scene, the object `chair' is found to be a \textit{relative} occlusion for target object based on dimensional relationship. Hence, the robot moves forward towards the `bottle', `table' and `chair'; avoids the `chair' and eventually, the target `cup' becomes visible. Hence, the approach was worked satisfactorily in simulation and real scenes. Online demo video: \url{https://youtu.be/Eps92qpYnRI}.
\begin{table}[h]
	\centering
	\begin{tabular}{|c c c c|} 
		\hline
		Player & Minimum & Maximum & Average \\ 
		\hline
		Human & 83 & 786 & 632 \\ 
		\hline
		Robot & - & - & 117 \\
		\hline
	\end{tabular}
	\caption{Comparison of time (in seconds) taken by humans and robot to perform same task with same input and control}
	\label{table:1}
\end{table}
\section{Conclusion}
In this paper, we presented an approach for a wheeled service robot to navigate in unknown scenes based only on ego RGB camera perception. The approach based on GeoSem map and rich semantic-based decision making was found to work satisfactorily in both simulated environment and real world deployment in indoor settings. Stream reasoning \cite{mukherjee2018system} \cite{mukherjee2013towards} \cite{banerjee2018system} can enrich the semantic processing of perception to make real time decisions. Future work will estimate depth from RGB view to enable dynamic obstacle avoidance as well an enhanced semantic scene understanding.

\bibliographystyle{IEEEtran}
\bibliography{root}

\begin{thebibliography}{10}
\providecommand{\url}[1]{#1}
\csname url@samestyle\endcsname
\providecommand{\newblock}{\relax}
\providecommand{\bibinfo}[2]{#2}
\providecommand{\BIBentrySTDinterwordspacing}{\spaceskip=0pt\relax}
\providecommand{\BIBentryALTinterwordstretchfactor}{4}
\providecommand{\BIBentryALTinterwordspacing}{\spaceskip=\fontdimen2\font plus
\BIBentryALTinterwordstretchfactor\fontdimen3\font minus
  \fontdimen4\font\relax}
\providecommand{\BIBforeignlanguage}[2]{{%
\expandafter\ifx\csname l@#1\endcsname\relax
\typeout{** WARNING: IEEEtran.bst: No hyphenation pattern has been}%
\typeout{** loaded for the language `#1'. Using the pattern for}%
\typeout{** the default language instead.}%
\else
\language=\csname l@#1\endcsname
\fi
#2}}
\providecommand{\BIBdecl}{\relax}
\BIBdecl

\bibitem{gireesh2022object}
N.~Gireesh and et. al., ``Object goal navigation using data regularized
  q-learning,'' \emph{arXiv preprint arXiv:2208.13009}, 2022.

\bibitem{kiran2022spatial}
D.~Kiran and et. al., ``Spatial relation graph and graph convolutional network
  for object goal navigation,'' \emph{preprint arXiv:2208.13031}, 2022.

\bibitem{tsintotas2022revisiting}
K.~A. Tsintotas and et. al., ``The revisiting problem in simultaneous
  localization and mapping: A survey on visual loop closure detection,''
  \emph{IEEE Transactions on Intelligent Transportation Systems}, 2022.

\bibitem{he2017mask}
K.~He, G.~Gkioxari, P.~Doll{\'a}r, and R.~Girshick, ``Mask r-cnn,'' in
  \emph{IEEE ICCV}, 2017, pp. 2961--2969.

\bibitem{redmon2018yolov3}
J.~Redmon and A.~Farhadi, ``Yolov3: An incremental improvement,'' \emph{arXiv
  preprint arXiv:1804.02767}, 2018.

\bibitem{han2018advanced}
J.~Han and et. al., ``Advanced deep-learning techniques for salient and
  category-specific object detection: a survey,'' \emph{IEEE Signal Processing
  Magazine}, vol.~35, no.~1, pp. 84--100, 2018.

\bibitem{gandon2018survey}
F.~Gandon, ``A survey of the first 20 years of research on semantic web and
  linked data,'' \emph{Ing{\'e}nierie des Syst{\`e}mes d'information}.

\bibitem{tenorth2017representations}
M.~Tenorth and et. al., ``Representations for robot knowledge in the knowrob
  framework,'' \emph{AI Journal}, vol. 247, pp. 151--169, 2017.

\bibitem{krishna2017visual}
R.~Krishna and et. al., ``Visual genome: Connecting language and vision using
  crowdsourced dense image annotations,'' \emph{International Journal of
  Computer Vision}, vol. 123, no.~1, pp. 32--73, 2017.

\bibitem{saputra2018visual}
M.~R.~U. Saputra, A.~Markham, and N.~Trigoni, ``Visual slam and structure from
  motion in dynamic environments: A survey,'' \emph{ACM Computing Surveys
  (CSUR)}, vol.~51, no.~2, pp. 1--36, 2018.

\bibitem{mukherjee2018system}
D.~Mukherjee, P.~Misra, and S.~Banerjee, ``System and a method for reasoning
  and running continuous queries over data streams,'' Jun.~5 2018, uS Patent
  9,990,403.

\bibitem{mukherjee2013towards}
D.~Mukherjee, S.~Banerjee, and P.~Misra, ``Towards efficient stream
  reasoning,'' in \emph{OTM ODBASE}.\hskip 1em plus 0.5em minus 0.4em\relax
  Springer, 2013, pp. 735--738.

\bibitem{banerjee2018system}
S.~Banerjee and D.~Mukherjee, ``System and method for executing a sparql
  query,'' Feb.~20 2018, uS Patent 9,898,502.

\end{thebibliography}

\end{document}